\documentclass[10pt,twocolumn,letterpaper]{article}

\usepackage{cvpr}
\usepackage{dsfont}
\usepackage{indentfirst}
 \usepackage{amsfonts}

\usepackage{graphicx}
\usepackage{amsmath}

\usepackage{amssymb}
\usepackage{booktabs}

\usepackage{float}
\usepackage{multirow}
\usepackage{diagbox}
\usepackage{balance}
\usepackage[linesnumbered,titlenumbered,ruled,vlined,commentsnumbered,noend]{algorithm2e}

\usepackage[T1]{fontenc}
\usepackage{xcolor}
\usepackage{epsfig}

\usepackage{enumitem}
\setenumerate[1]{itemsep=0pt,partopsep=0pt,parsep=\parskip,topsep=5pt}
\setitemize[1]{itemsep=0pt,partopsep=0pt,parsep=\parskip,topsep=5pt}
\setdescription{itemsep=0pt,partopsep=0pt,parsep=\parskip,topsep=5pt}

\usepackage{moresize}
\usepackage{epstopdf}
\usepackage{array}

\usepackage{makecell}
\usepackage{bm}
\usepackage{dsfont}

\usepackage{amssymb}
\usepackage{sidecap}
\usepackage{colortbl}

\usepackage[font=footnotesize,labelfont=bf]{caption}
\definecolor{citecolor}{RGB}{65,105,225}
\usepackage[breaklinks=true,colorlinks,citecolor=citecolor,bookmarks=false]{hyperref}

\usepackage{dsfont}
\usepackage{xspace}

\def\argmin{\operatornamewithlimits{arg\,min}}

\definecolor{dg}{rgb}{0,0.694,0.298}
\definecolor{purple}{rgb}{0.4,0.176,0.569}
\definecolor{royalblue}{RGB}{65,105,225}

\usepackage{pifont}
\newcommand{\figref}[1]{Fig.~\ref{#1}}
\newcommand{\reqref}[1]{Eq.~\eqref{#1}}
\newcommand{\secref}[1]{Sec.~\ref{#1}}

\usepackage{xspace}
\makeatletter
\DeclareRobustCommand\onedot{\futurelet\@let@token\@onedot}
\def\@onedot{\ifx\@let@token.\else.\null\fi\xspace}
\def\eg{\emph{e.g}\onedot} 
\def\ie{\emph{i.e}\onedot}

\makeatother

\definecolor{americanrose}{rgb}{1.0, 0.01, 0.24}

\newcommand{\topone}[1]{\textbf{\textcolor{red}{#1}}}

\usepackage[capitalize]{cleveref}
\crefname{section}{Sec.}{Secs.}
\Crefname{section}{Section}{Sections}
\Crefname{table}{Table}{Tables}
\crefname{table}{Tab.}{Tabs.}

\title{\textsc{CosalPure}: Learning Concept from Group Images for \\ Robust Co-Saliency Detection}

\author{Jiayi Zhu\textsuperscript{1},\quad Qing Guo\textsuperscript{3}$^\dagger$, \quad Felix Juefei-Xu\textsuperscript{2},\quad Yihao Huang\textsuperscript{4}, \quad Yang Liu\textsuperscript{4}, \quad Geguang Pu\textsuperscript{1,5}$^\dagger$\\
~\\
\textsuperscript{1} East China Normal University, China \quad
\textsuperscript{2} New York University, USA \\
\textsuperscript{3} IHPC \& CFAR, Agency for Science, Technology and Research, Singapore \\
\textsuperscript{4} Nanyang Technological University, Singapore \\
\textsuperscript{5} Shanghai Industrial Control Safety Innovation Tech. Co., Ltd, China
}

\begin{document}

\twocolumn[{
\renewcommand\twocolumn[1][]{#1}
\maketitle
\begin{center}
    \captionsetup{type=figure}
    \vspace{-10pt}
    \includegraphics[width=\textwidth]{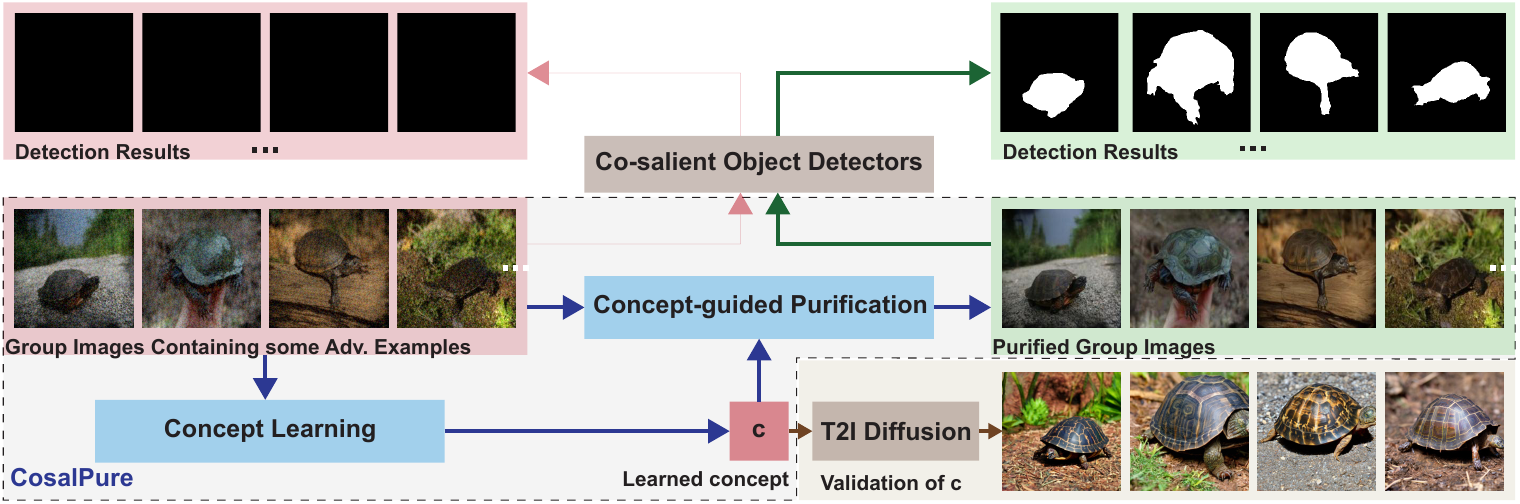}
    \captionof{figure}{Examples of our method \textsc{CosalPure} and comparative results before and after purification.
    \textsc{CosalPure} comprises two modules: group-image concept learning and concept-guided purification.
    Firstly, the concept learning module inputs a group of images that contain some adversarial cases and obtain their shared co-salient semantic information (\ie, the learned concept), denoted as \textbf{c}.
    We can validate the effectiveness of the learned \textbf{c} through the visualization via a text-to-image (T2I) diffusion model.
    Secondly, steered by the previously learned concept, we employ certain diffusion generation techniques to purify the entire group of images.
    Before our purification, the co-salient object detection results are poor, but after purification, the detection results are satisfactory. Please enlarge to see more details.
}
    \label{fig:teaser}
\end{center}
}]
\renewcommand{\thefootnote}{}
\footnotetext{$^\dagger$ Geguang Pu (ggpu@sei.ecnu.edu.cn) and Qing Guo (tsingqguo@ieee.org) are corresponding authors.}

\begin{abstract}
Co-salient object detection (CoSOD) aims to identify the common and salient (usually in the foreground) regions across a given group of images. 
Although achieving significant progress, state-of-the-art CoSODs could be easily affected by some adversarial perturbations, leading to substantial accuracy reduction.
The adversarial perturbations can mislead CoSODs but do not change the high-level semantic information (\eg, concept) of the co-salient objects.
In this paper, we propose a novel robustness enhancement framework by first learning the concept of the co-salient objects based on the input group images and then leveraging this concept to purify adversarial perturbations, which are subsequently fed to CoSODs for robustness enhancement. 
Specifically, we propose \textsc{CosalPure} containing two modules, \ie, group-image concept learning and concept-guided diffusion purification. 
For the first module, we adopt a pre-trained text-to-image diffusion model to learn the concept of co-salient objects within group images where the learned concept is robust to adversarial examples.
For the second module, we map the adversarial image to the latent space and then perform diffusion generation by embedding the learned concept into the noise prediction function as an extra condition. 
Our method can effectively alleviate the influence of the SOTA adversarial attack containing different adversarial patterns, including exposure and noise.
The extensive results demonstrate that our method could enhance the robustness of CoSODs significantly.
The project is available at \href{https://v1len.github.io/CosalPure/}{https://v1len.github.io/CosalPure/}.
\end{abstract}


\section{Introduction}



Co-salient object detection (CoSOD) plays a pivotal role in visual information analysis, aiming to identify and accentuate common and salient objects across a set of images \cite{zhang2018review}. This study area, crucial for applications like image segmentation and object recognition, has witnessed considerable advancement with the advent of neural network-based methodologies. These methods excel in discerning shared saliency cues among images, offering a significant leap over traditional saliency detections \cite{liu2019simple}. However, their robustness is severely tested under adverse conditions, such as adversarial attacks and various image common corruption, including but not limited to motion blur \cite{hendrycks2019robustness}.


The susceptibility of CoSOD methods to adversarial perturbations, such as those introduced by Jadena \cite{gao2022can}, poses a significant challenge. These perturbations, while not altering the high-level semantic information of images, can drastically reduce the accuracy of co-salient object detection. The disparity between the corrupted image's saliency map and the ground truth, as a result of these attacks, highlights a critical vulnerability in current CoSOD approaches.

%
%

Currently, there are indeed methods aimed at defending against adversarial attacks, such as DiffPure, which employs a noise addition and denoising strategy to eliminate perturbations. 
However, when restoring the image, DiffPure does not take into account the identity of the object within the image (\ie, without object-specific information). As a result, the restored images produced by DiffPure may contain artifacts that are artificially generated. These artifacts will affect the detection results of the CoSOD method.

To fill this gap, this work introduces a novel robustness enhancement framework, \textsc{CosalPure}.
The intuitive idea is to first learn a concept from the group images and then use it to guide the data purification based on text-to-image (T2I) diffusion.
The `concept' means the high-level semantic information of co-salient objects in the group images and falls within the text's latent space.
Specifically, this innovative approach comprises two meticulously designed modules: group-image concept learning and concept-guided diffusion purification. The first module focuses on learning the concept of co-salient objects from group images, demonstrating robustness and resilience to adversarial examples. The second module strategically maps adversarial images into a latent space, following which diffusion generation techniques, steered by the previously learned concept, are employed to purify these images effectively.


As shown in the left panel of \figref{fig:teaser}, when a group of images containing some adversarial examples is passed into co-salient object detectors, the detection results are poor.
We first apply the concept learning module to obtain the shared co-salient semantic information \textbf{c} (\ie, the learned concept).
The bottom right corner of \figref{fig:teaser} shows the visualization results of \textbf{c} via a T2I diffusion model.
It is evident that the semantic information in the visualized images aligns with the original group images, demonstrating the effectiveness of the concept learning module.
Secondly, we utilize the just-proven effective learned concept \textbf{c} to guide the purification.
From the right panel of \figref{fig:teaser}, we can observe that the purified group images via \textsc{CosalPure} exhibit satisfactory performance in the CoSOD task.
%

Extensive experimental results substantiate the effectiveness of \textsc{CosalPure}. \textsc{CosalPure} stands as a robust, concept-driven solution, paving the way for more reliable and accurate co-salient object detection in an era where image manipulation and corruption are increasingly prevalent.


\section{Related Work}

\noindent\textbf{Co-salient object detection.}
Different from single-image saliency detection \cite{achanta2009frequency,liu2019simple,cheng2014global,li2016deep,wang2016saliency,wang2021salient}, the goal of co-saliency detection is to detect common salient objects in a group of images \cite{fan2021re,jiang2019unified,li2019detecting,zha2020robust,zhang2020adaptive,zhang2020gradient}, evolving from early feature-based approaches to sophisticated deep learning and semantic-driven methods.
Deciphering correspondences among co-salient objects across multiple images is pivotal for co-saliency detection. This challenge can be effectively tackled through optimization-based methods \cite{cao2014self,li2014repairing}, machine learning-based models \cite{cheng2014salientshape,zhang2015self}, and deep neural networks \cite{zha2020robust,zhang2020adaptive,zhang2020gradient}.
GICD \cite{zhang2020gradient} employs a gradient-induced mechanism that pays more attention to discriminative convolutional kernels which helps to locate the co-salient regions.
GCAGC \cite{zhang2020adaptive} presents an adaptive graph convolutional network with attention graph clustering for co-saliency detection.

\noindent\textbf{Adversarial attack for co-salient object detection.}
Jadena \cite{gao2022can} is an adversarial attack that jointly tunes the exposure and additive perturbations, which can drastically reduce the accuracy of co-salient object detection.


\noindent\textbf{Text-to-image diffusion generation model.}
The popularity of Text-to-Image (T2I) generation \cite{zhang2023text} is propelled by diffusion models \cite{croitoru2023diffusion,ho2020denoising,rombach2022high}, necessitating training on extensive text and image paired datasets like LAION-5B \cite{schuhmann2022laion}. 
The adeptly trained model demonstrates proficiency in generating diverse and lifelike images based on user-specific input text prompts, realizing T2I generation. 
T2I personalization \cite{gal2022image,ruiz2023dreambooth} is geared towards steering a diffusion-based T2I model to generate innovative concepts.

\noindent\textbf{Diffusion-based image purification methods.}
DiffPure \cite{nie2022diffusion} is a notable approach in the field of image processing, specifically designed to enhance the robustness of images against adversarial attacks. 
It employs a strategy of introducing controlled noise via the forward stochastic differential equation (SDE) \cite{meng2021sdedit} and subsequently denoising the image via the reverse SDE to counteract adversarial perturbations. 
While effective in reducing these perturbations, it is noteworthy that DiffPure does not explicitly consider object semantics during the image restoration process.
Diffusion-Driven Adaptation (DDA) \cite{gao2022back} is a test-time adaptation method that improves model accuracy on shifted target data by updating inputs through a diffusion model \cite{ho2020denoising}, effectively avoiding domain-wise re-training.


\section{Preliminaries and Motivation}


\subsection{Co-salient Object Detection (CoSOD)}
We have a group of images $\mathcal{I} = \{\mathbf{I}_i \in \mathds{R}^{H\times W\times 3}\}_{i=1}^{N}$ that contain $N$ images, and these images have common salient objects.
We denote a CoSOD method as $\textsc{CoSOD}(\cdot)$, taking $\mathcal{I}$ as input and predicting $N$ salient maps,
\begin{align}
    \mathcal{S} = \{\mathbf{S}_i\}_{i=1}^{N} = \textsc{CoSOD}(\mathcal{I})    \label{eq:cosal},
\end{align}
%
where $\mathbf{S}_i\in \mathds{R}^{H\times W}$ is a binary map (\ie, the saliency map) indicating the salient region of the $i$-th image $\mathbf{I}_i$.
%
%
We show an example of co-saliency detection results in \figref{fig:teaser}.

\subsection{Robust Issues of CoSOD}

However, at times, part of the acquired images may be of low quality (\ie, corrupted by some degradation), which will affect the robustness of CoSOD methods.
In particular, \cite{gao2022can} proposes the joint adversarial noise and exposure attack that can reduce the detection accuracy of state-of-the-art CoSODs significantly.
To be specific, within the entire group, there are $M$ images that have been added adversarial perturbations, denoted as $\{\mathbf{I}_{j}^{'}\}_{j=1}^{M}$, while the remaining are considered as clean images $\{\mathbf{I}_{k}\}_{k=1}^{N-M}$.
In this scenario, we can reformulate \reqref{eq:cosal} as
%
\begin{align}
    \mathcal{S}^{'} = \{\mathbf{S}_i^{'}\}_{i=1}^{N} = \text{CoSOD}(\{\mathbf{I}_{j}^{'}\}_{j=1}^{M} \cup \{\mathbf{I}_{k}\}_{k=1}^{N-M})
    \label{eq:cosal_reformulate}.
\end{align}
%
The difference between $\mathcal{S}$ from \reqref{eq:cosal} and $\mathcal{S}'$ indicates the robustness of the CoSOD method. 
Previous research findings indicate that existing CoSOD methods are susceptible to the influence of anomalous data (\eg, adversarial noise and exposure) \cite{gao2022can}.
Note that the degraded images (\eg, $\{\mathbf{I}_{j}^{'}\}_{j=1}^{M}$) may not only affect the saliency maps of themselves but also impact the saliency maps of clean images.


Therefore, to enhance the robustness of CoSOD methods, defense methods should be developed. However, few works are focusing on this direction.
A typical defense method is to purify the input images to remove the effects of degradations.
In the following, we study the SOTA purification method, \ie, DiffPure \cite{nie2022diffusion}, to enhance the robustness and show that enhancing the robustness of CoSOD is a non-trivial task and new technologies should be developed.


\subsection{DiffPure and Challenges}
%
\begin{figure}
\centering
\includegraphics[width=1\linewidth]{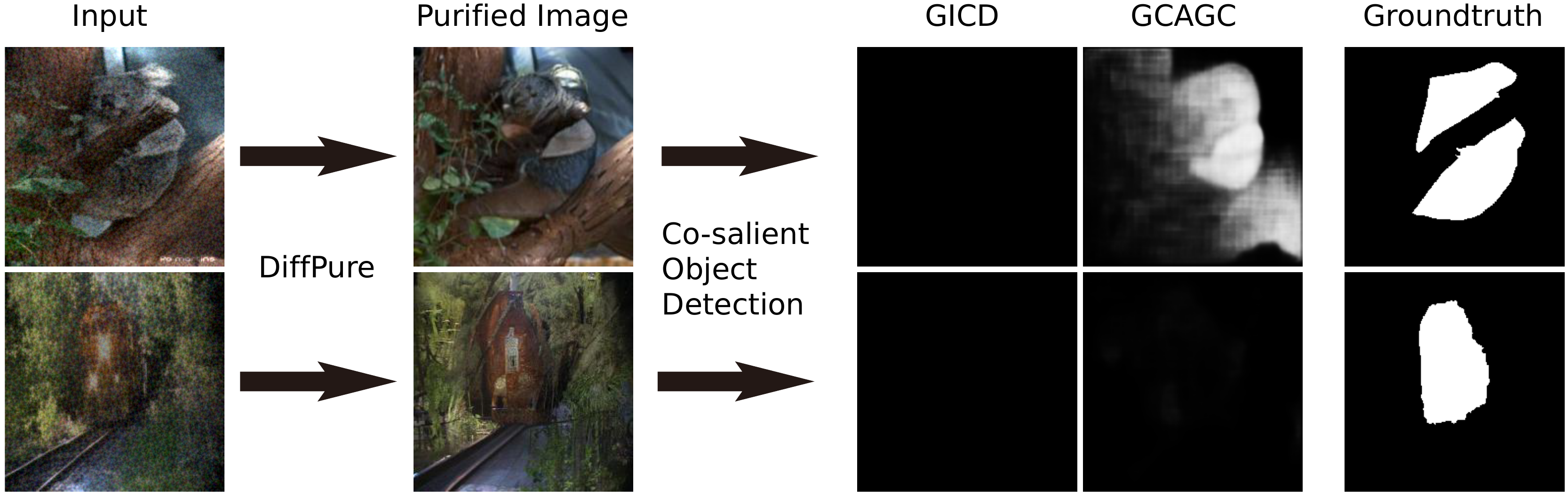}
\caption{CoSOD results for DiffPure.The input images are under the attack method \cite{gao2022can}. Processed by DiffPure \cite{nie2022diffusion}, the purified images perform inferior in the CoSOD task together with their respective group images.}
\label{fig:DiffPure_CoSOD}
\vspace{-10pt}
\end{figure}
A highly intuitive approach is to perform image reconstruction on input images, hoping to remove degradations.
As an existing method for image reconstruction, DiffPure \cite{nie2022diffusion} can remove adversarial perturbations by applying forward diffusion followed by a reverse generative process.
However, DiffPure is limited in its ability to address adversarial additive perturbations. It presents limited capabilities for handling other degradations like adversarial exposure.
%
\figref{fig:DiffPure_CoSOD} illustrates two cases, with the upper case depicting a koala and the lower case representing a train.
The input images for both cases are under the attack method \cite{gao2022can}. The images processed through DiffPure visually eliminate perturbations.
However, when the purified images underwent co-salient object detection together with the images within their respective groups, the detection results are inferior.
\figref{fig:DiffPure_CoSOD} illustrates that the DiffPure cannot enhance the robustness of CoSOD under the attack method \cite{gao2022can}.
%

%
%
We tend to design a more effective purification method.
DiffPure is specifically designed against adversarial attacks for image classification and neglects the specific properties of the CoSOD task: 
\ding{182} Only partial images within the group are attacked, and the clean images contain rich complementary information, which could help enhance the robustness.
\ding{183} Although the adversarial patterns may affect the semantic features of images, the fact group images contain co-salient objects has not changed. How to utilize such a property should be carefully studied.



\section{Methodology: \textsc{CosalPure}}

\begin{figure*}[t]
\centering
\includegraphics[width=1\linewidth]{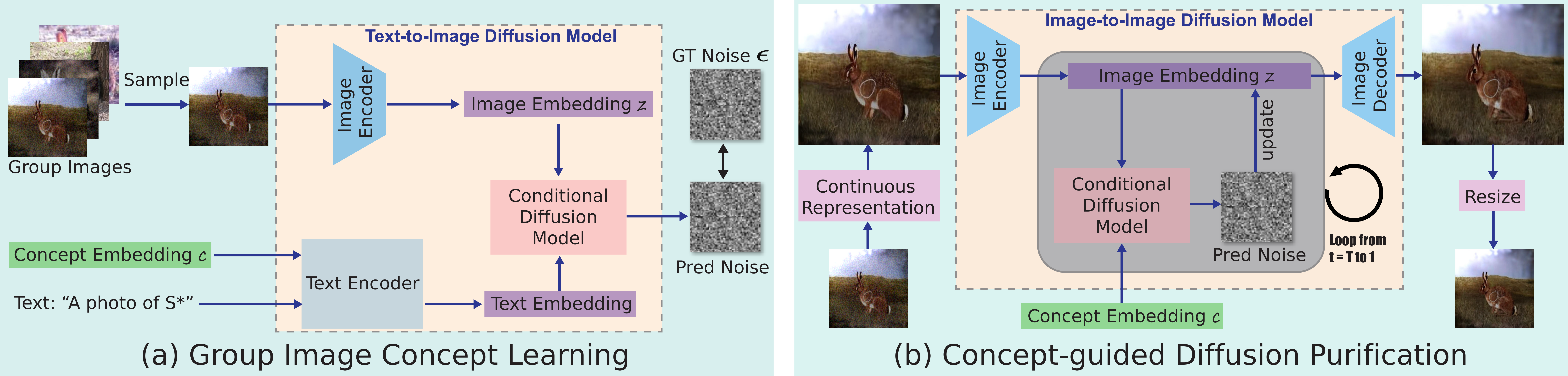}
\caption{Overview of \textsc{CosalPure}.
The details of (a) are in \secref{subsec:concept_learn}, while the details of (b) are in \secref{subsec:concept_inversion}.
}
\label{fig:overview}
\end{figure*}

\subsection{Overview} 
\label{subsec:overview}

Beyond DiffPure \cite{nie2022diffusion}, we propose to learn the concept of co-salient objects from the group images and leverage it to guide the purification.
Specifically, given group images $\mathcal{I}' = \{\mathbf{I}_{j}^{'}\}_{j=1}^{M} \cup \{\mathbf{I}_{k}\}_{k=1}^{N-M} $ that contains $M$ degraded images and $N-M$ clean images, we first learn the concept from $\mathcal{I}'$ via the recent developed textual inversion method. 
The learned concept is a token and lies in the latent space of texts.
We name it as `concept' since we can use it to generate new images containing the `concept'. 
Note that the number of the degraded images (\ie, $M$) is unknown during application.
We denote the concept of learning as
\begin{align} \label{eq:conceptlearn}
    \mathbf{c} = \text{ConceptLearn}(\mathcal{I}'),
\end{align}
and we detail the whole process in \secref{subsec:concept_learn}.

After obtaining the concept, we aim to leverage it for purification by
\begin{align} \label{eq:cosalpure}
    \hat{\mathbf{I}} = \text{ConceptPure}(\mathbf{c},\mathbf{I}), \mathbf{I}\in \mathcal{I}',
\end{align}
where the image $\hat{\mathbf{I}}$ is the purified image of $\mathbf{I}$ that may be a clean image or a perturbed image. We detail the concept-guided diffusion purification in \secref{subsec:concept_inversion}.
For each image in $\mathcal{I}'$, we can handle it via \reqref{eq:cosalpure} and get a novel group denoted as $\hat{\mathcal{I}}$. 
Then, we feed $\hat{\mathcal{I}}$ to CoSOD methods to see whether their robustness is enhanced or not.

The core idea is valid based on a critical assumption: the perturbed images in $\mathcal{I}'$ do not affect the concept learning. 
We detail this in the \secref{subsec:concept_learn}.

\begin{figure}
\centering
\includegraphics[width=1\linewidth]{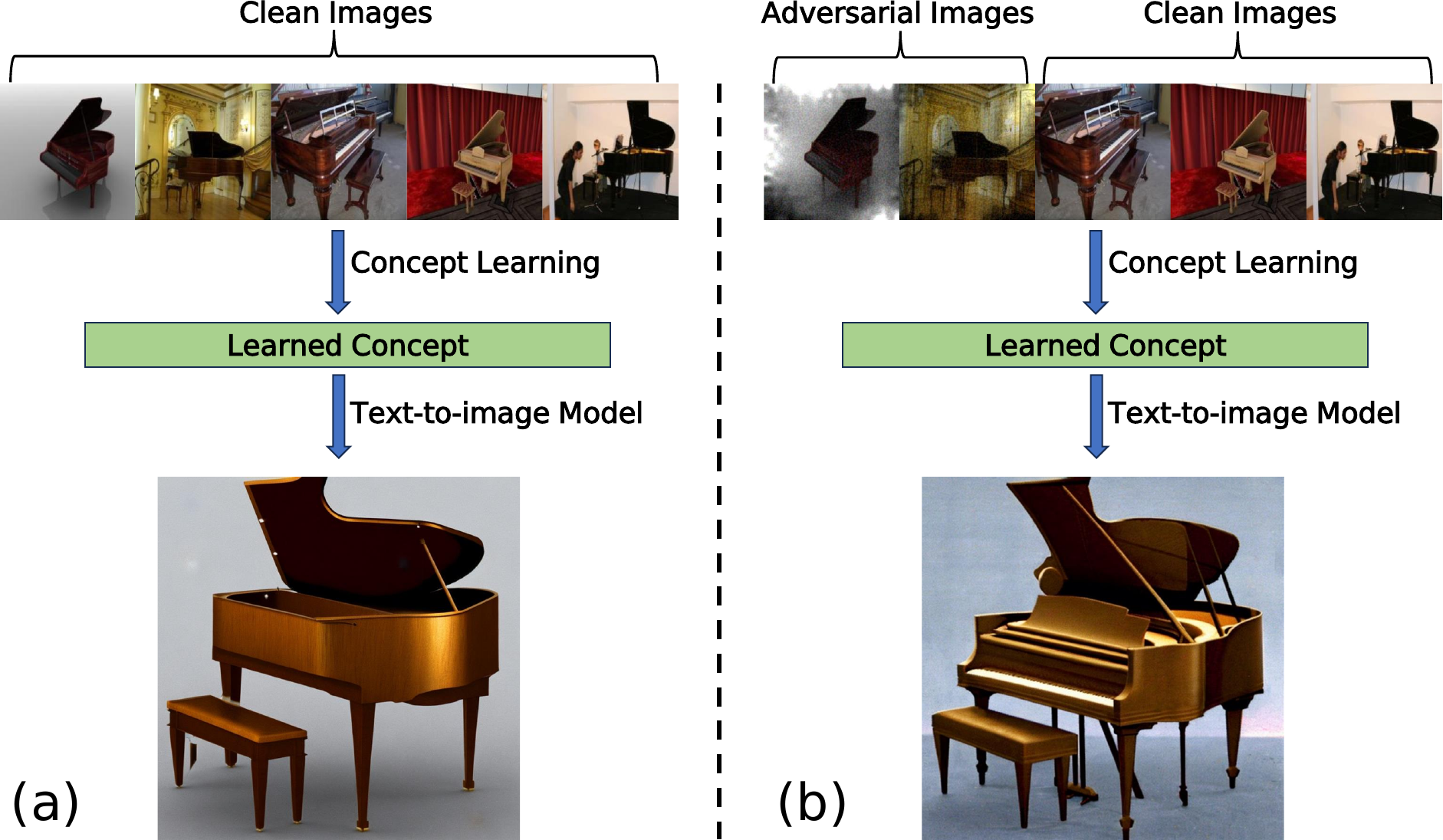}
\caption{Demonstration of the effectiveness of concept learning. (a) Five clean images are utilized for concept learning, and the learned concept can be reconstructed into an image through a pre-trained text-to-image model. (b) The first two images are attacked by Jadena \cite{gao2022can} while the subsequent three images are clean, and the learned concept can also be reconstructed into a high-quality image. (a) and (b) use the same random seed.}
\label{fig:concept_learning_figure}
\end{figure}

\subsection{Group-Image Concept Learning}
\label{subsec:concept_learn}

In this section, we introduce the detail of group-image concept learning (\ie, Eq~\eqref{eq:conceptlearn}), which tends to utilize a group of input images for learning the text-aligned embedding of common objects they have, as shown in \figref{fig:overview} (a).
We denote this process as
    $\textbf{c} =\text{ConceptLearn}\left(\{\mathbf{I}_{j}^{'}\}_{j=1}^{M} \cup \{\mathbf{I}_{k}\}_{k=1}^{N-M}\right)$
where $\textbf{c}$ represents a token aligned with the texts' latent space and represents the semantic information of common objects.

To this end, we formulate group-image concept learning as the personalizing text-to-image problem \cite{gal2022image,ruiz2023dreambooth} to enable text-to-image (T2I) diffusion models to rapidly swift new concept acquisition.

\textbf{Text-to-Image Diffusion Model.}
We introduce the architecture and procedure of a classical T2I diffusion model \cite{rombach2022high}. It consists of three core modules: (1) image autoencoder, (2) text encoder, (3) and conditional diffusion model.
The \textit{\underline{image autoencoder}} module has two submodules: an encoder $\mathcal{E}$ and a decoder $\mathcal{D}$. It serves a dual purpose, where the encoder maps an input image $\mathbf{X}$ to a low-dimensional latent space with $\textbf{z} = \mathcal{E}(\mathbf{X})$, while the decoder transforms the latent representation back into the image space with $\textit{D}(\mathcal{E}(\mathbf{X})) \approx \mathbf{X}$.
The \textit{\underline{text encoder}} $\Gamma$ firstly processes a text $\textbf{y}$ by tokenizing it and secondly translates it into a latent space text embedding 
$\Gamma(\textbf{y})$.
The \textit{\underline{conditional diffusion model}} $\epsilon_\theta$ takes the time step $t$, the noisy latent $\textbf{z}_t$ at $t$-th time step and the text embedding $\Gamma(\textbf{y})$ as input to predict the noise added on $\textbf{z}_t$, denoted as $\epsilon_\theta(\textbf{z}_t,t,\Gamma(\textbf{y}))$.

%
Given a pre-trained T2I diffusion model and group images $\mathcal{I}'=\{\text{I}_{j}^{'}\}_{j=1}^{M} \cup \{\text{I}_{k}\}_{k=1}^{N-M}$ used for CoSOD task, we aim to learn a concept of the common object within $\mathcal{I}'$ by
%
\begin{align}
    %
    \textbf{c} = & \argmin\limits _{\textbf{c}^*} \mathbb{E}_{\mathbf{X}\in \mathcal{I}', \textbf{z}\in \mathcal{E} 
    (\mathbf{X}), \mathbf{y},\epsilon\in \mathcal{N}(0,1),t} ( \nonumber \\
    & \|\epsilon_\theta(\mathbf{z}_t,t, \Upsilon(\Gamma(\mathbf{y}),\textbf{c}^*))-\epsilon\|_2^2),
    \label{eq:argmin_concept}
\end{align}
%
where $\mathbf{y}$ is a fixed text (\ie, `a photo of $S^*$') and the function $\Upsilon(\Gamma(\mathbf{y}),\textbf{c}^*)$ is to replace the token of `$S^*$' within $\Gamma(\mathbf{y})$ with $\textbf{c}^*$.  
Intuitively, \reqref{eq:argmin_concept} forces the concept $\mathbf{c}^*$ to represent the co-salient objects within group images and also lies in the text latent space corresponding to the text `$S^*$'. 
After obtaining $\mathbf{c}$, we can embed `$S^*$' into other texts to generate new images via the T2I diffusion model. 
For example, in \figref{fig:concept_learning_figure} (b), we learn a concept of the co-salient objects (\ie, piano), which corresponds to the text `$S^*$'. 
Then, we feed a text (\eg, `a photo of $S^*$') to the T2I model that generates an image containing the object, which means that the learned concept represents the salient objects in $\mathcal{I}'$ very well.

\textbf{Robustness of concept learning.} 
It is obvious that the above method naturally aligns with our objective since we can exploit it to obtain the common semantic content among the group images for the CoSOD task. 
The key problem is whether the concept learning would be affected by degradations like adversarial perturbation in the group images. 
We conduct an empirical study to validate this.
Specifically, given a group of clean images (\ie, $\mathcal{I}$), we use it to learn a concept via \reqref{eq:argmin_concept}. 
Meanwhile, we conduct the adversarial CoSOD attack \cite{gao2022can} on two images within $\mathcal{I}$ and form a new group $\mathcal{I}'$.
With $\mathcal{I}'$, we learn another concept via \reqref{eq:argmin_concept}.
Then, we can leverage the two learned concepts to generate images based on the same text prompt.
As shown in \figref{fig:concept_learning_figure}, we find that the two generated images based on two concepts are similar, demonstrating that the adversarial examples have limited influence on concept learning.
This inspires us to leverage the learned concept to purify the adversarial examples.


\subsection{Concept-guided Diffusion Purification}
\label{subsec:concept_inversion}


We propose reconstructing the group images based on the learned concept $\textbf{c}$ to eliminate the potential adversarial patterns as shown in \figref{fig:overview} (b).
%
%
Considering the advantage of the continuous representation \cite{chen2021learning, ho2022disco} in the smooth image reconstruction and its ability to remove perturbations, we employ a continuous representation module for the initial processing of the input image $\mathbf{X}$, denoted as $\tilde{\mathbf{X}} = \textsc{CR}(\mathbf{X})$. 
This module not only somewhat denoises $\mathbf{X}$ but also addresses the issue of pre-trained CoSOD models designed for specific resolutions that do not match the input resolution of our employed diffusion model.
Subsequently, the encoder of the image autoencoder module maps $\tilde{\mathbf{X}}$ to the latent space $\textbf{z}_0 = \mathcal{E}(\tilde{\mathbf{X}})$. 
The following procedure is based on the diffusion pipeline and needs two procedures: forward process and reverse process. 
The forward diffusion process is a fixed Markov chain that iteratively adds a Gaussian noise to the latent $\textbf{z}_0$ over $T$ timesteps, obtaining a sequence of noised images $\textbf{z}_1$, $\textbf{z}_2$, $\cdots$, $\textbf{z}_{T}$.
In each step of the forward progress, the latent at time step $t\in [1, T]$ is updated by
\begin{align}
    \mathbf{z}_t = a_t \mathbf{z}_{t-1} + b_t\epsilon_t, \epsilon_t \sim \mathcal{N}(\mathbf{0}, \mathbf{I}),
    \label{eq:forward}
\end{align}
where $a_t$ and $b_t$ are coefficients and $\mathcal{N}(0, \mathbf{I})$ represents the standard Gaussian distribution.
By superimposing time steps from $t=1$ to $T$, \reqref{eq:forward} can be simplified to 
\begin{align}
    \mathbf{z}_t = \sqrt{\overline{\alpha}_t}\textbf{z}_{0} + \sqrt{1-\overline{\alpha}_t}\epsilon_t,~ \epsilon_t \sim \mathcal{N}(0, \mathbf{I}),
    \label{eq:forward_simplified}
\end{align}
where we have $a_t^2~+~b_t^2=1$, and $\alpha_t = a_t^2$, $\overline{\alpha}_t = \prod_{\tau=1}^{t}{\alpha_\tau}$. As we set the time step as $T$, the complete forward process can be expressed as 
\begin{align}
   \mathbf{z}_{T} \sim \text{q}(\mathbf{z}_{1:T} | \mathbf{z}_0) = \prod_{t=1}^{T}{\text{q}(\textbf{z}_{t} | \mathbf{z}_{t-1})}.
    \label{eq:forward_simplified_probability_complete}
\end{align}

For the reverse process, it iteratively removes the noise to generate an image in $T$ timesteps. 
Unlike doing the reverse process directly, our method incorporates the obtained semantic embedding $\textbf{c}$ as additional object information into the pipeline.
%
%
Then, we can start from $\textbf{z}_T$ (alternatively called $\hat{\mathbf{z}}_{T}$) and progressively predict 
\begin{align}
    \epsilon_{t-1} = \epsilon_\theta(\hat{\mathbf{z}}_t,t,\mathbf{c}),
    \label{eq:pred_eplison}
\end{align}
and obtain the latent at time step $t-1$ via
%
\begin{align}
\hat{\mathbf{z}}_{t-1} = & \frac{\sqrt{\bar{\alpha }_{t-1}}(1-\alpha_t)}{1-\bar{\alpha}_t}\tilde{\mathbf{z}}_0 \nonumber\\ 
 + & \frac{\sqrt{\alpha }_{t}(1-\bar{\alpha}_{t-1})}{1-\bar{\alpha}_t}\hat{\mathbf{z}}_t+\sigma_t \xi,
    \label{eq:pred_zt-1}
\end{align}
%
with
\begin{align}
\tilde{\textbf{z}}_0 = \frac{\hat{\textbf{z}}_t - \sqrt{1-\bar{\alpha }_t } \epsilon_{t-1} }{\sqrt{\bar{\alpha}_t}},
    \label{eq:pred_z0}
\end{align}
%
where $\sigma_t^2 = \frac{(1-\alpha_t )(1-\overline{\alpha }_{t-1})}{1-\overline{\alpha }_{t}} $ and $\xi \sim \mathcal{N}(\textbf{0}, \mathbf{I})$ according to the sample process of DDPM \cite{ho2020denoising}. We can directly obtain the reconstructed image $\hat{\textbf{x}}$ by using the decoder with formula $\hat{\textbf{x}} = \mathcal{D}(\hat{\textbf{z}}_0)$.

\begin{figure}[t]
\centering
\includegraphics[width=1\linewidth]{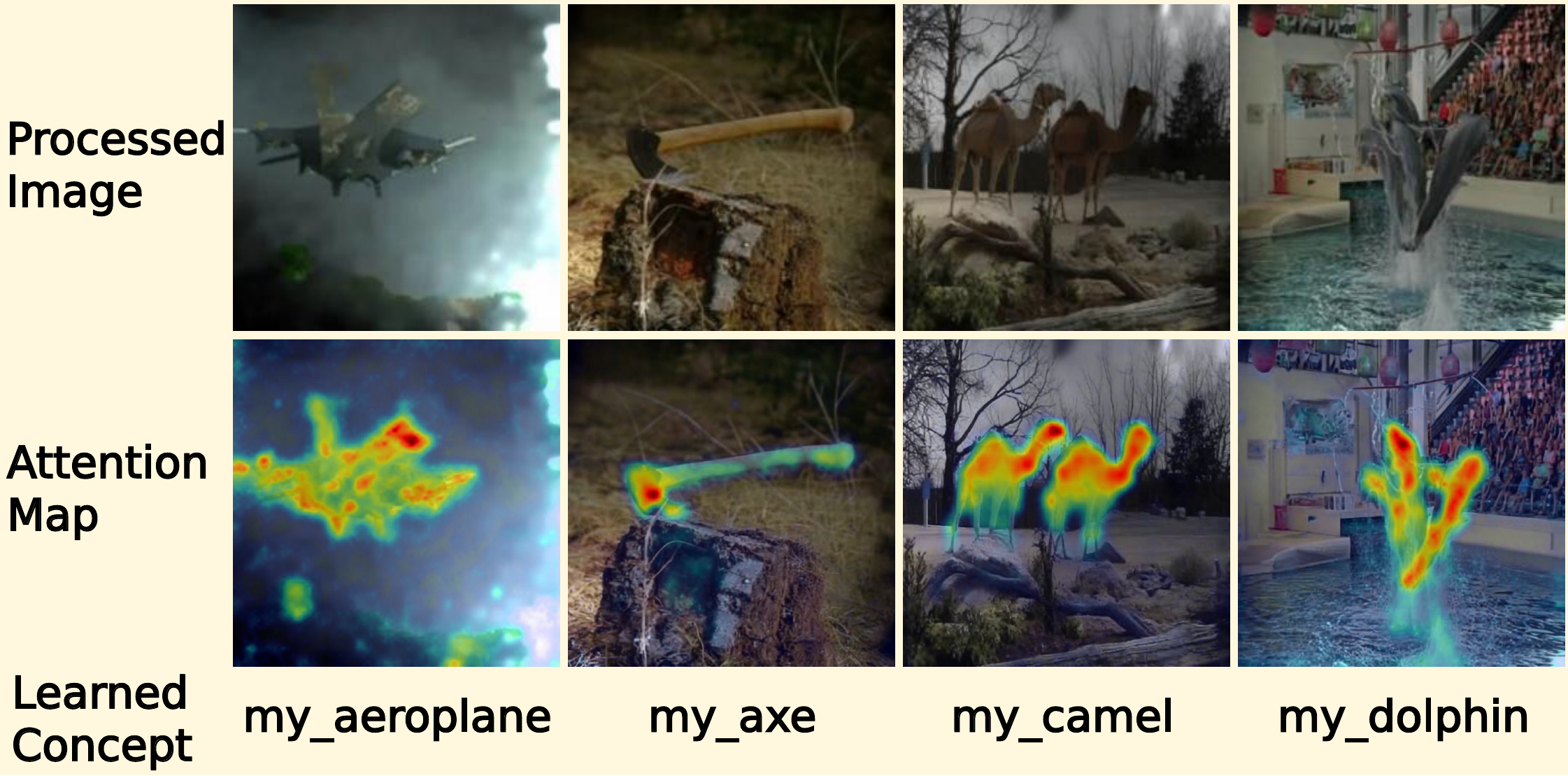}
\caption{Attention maps for learned concepts on processed images.}
\label{fig:attention}
\vspace{-10pt}
\end{figure}
To confirm that the concept learned by \textsc{CosalPure} is applied accurately in the image reconstruction, we employ DAAM \cite{tang2022daam} to establish attention maps for the learned concepts on processed images as shown in \figref{fig:attention}.
In each case, the attention map of the semantic embedding $\textbf{c}$ (\ie, the learned concept) aligns well with the object itself in the image, indicating the effectiveness of \textsc{CosalPure}.

%
%

\begin{table*}[ht]
\caption{Co-saliency detection performance.
“Source-Only” means the group of images before processing, including 50\% adversarial images and 50\% clean images.
We highlight the top results of each CoSOD method and each dataset in \topone{red}.}
\vspace{-5pt}
    \centering
    \label{Table:baselines}
    \setlength{\tabcolsep}{8pt}
    \resizebox{1\linewidth}{!}{

\begin{tabular}{ll|cccc|cccc|cccc}
\hline 
\multirow{2}{*}{} & \multirow{2}{*}{} & \multicolumn{4}{c|}{GICD} & \multicolumn{4}{c|}{GCAGC} & \multicolumn{4}{c}{PoolNet}\tabularnewline
 &  & SR $\uparrow$  & AP $\uparrow$ & $F_\beta$ $\uparrow$ & MAE $\downarrow$  &  SR $\uparrow$  & AP $\uparrow$ & $F_\beta$ $\uparrow$ & MAE $\downarrow$  &  SR $\uparrow$  & AP $\uparrow$ & $F_\beta$ $\uparrow$ & MAE $\downarrow$ \tabularnewline
\hline 
\hline 
\multirow{4}{*}{\rotatebox{90}{Cosal2015}} & Source-Only & 0.3493 & 0.7306  & 0.4038 & 0.1676 & 0.5285  & 0.7853 & 0.6302 & 0.1570 & 0.5677  & 0.7425 & 0.6095 & 0.1276\tabularnewline
 & DiffPure & 0.4595  & 0.7478  & 0.5118  & 0.1444  & 0.4987  & 0.6998  & 0.5901  & 0.2162  & 0.6327  & 0.7779  & 0.6714  & 0.1181 \tabularnewline
 & DDA & 0.4565  & 0.7579  & 0.5158  & 0.1469  & 0.5955  & \topone{0.7928}  & \topone{0.6774}  & \topone{0.1542}  & 0.6233  & 0.7863  & 0.6691  & 0.1181 \tabularnewline
 & \textsc{CosalPure} & \topone{0.5602}  & \topone{0.7898}  & \topone{0.6177}  & \topone{0.1296}  & \topone{0.5975}  & 0.7449  & 0.6521  & 0.2063  & \topone{0.6908} & \topone{0.8268}  & \topone{0.7258} & \topone{0.1086} \tabularnewline
\hline 
\hline 
\multirow{4}{*}{\rotatebox{90}{iCoseg}} & Source-Only & 0.4012  & 0.7269  & 0.5063  & 0.1420  & 0.6469  & 0.8237  & 0.7173  & \topone{0.1146}  & 0.5847  & 0.8116 & 0.6472  & 0.1057 \tabularnewline
 & DiffPure & 0.4447 & 0.7291  & 0.5503  & 0.1269  & 0.6609  & 0.8043 & 0.7051  & 0.1257 & 0.6796  & 0.8328  & 0.7144  & 0.0905 \tabularnewline
 & DDA & 0.4665  & 0.7519  & 0.5948  & 0.1280  & 0.6982  & \topone{0.8257}  & \topone{0.7390}  & 0.1235 & 0.6578  & 0.8483  & 0.7179  & 0.0940 \tabularnewline
 & \textsc{CosalPure} & \topone{0.5396}  & \topone{0.7611}  & \topone{0.6329}  & \topone{0.1208}  & \topone{0.7060}  & 0.8052  & 0.7265  & 0.1413  & \topone{0.7278}  & \topone{0.8730}  & \topone{0.7577}  & \topone{0.0850} \tabularnewline
\hline 
\hline 
\multirow{4}{*}{\rotatebox{90}{CoSOD3k}} & Source-Only & 0.3281  & 0.6988  & 0.4003  & 0.1439  & 0.4445  & 0.7325  & 0.5702  & 0.1376  & 0.4466  & 0.6606  & 0.5255  & 0.1386 \tabularnewline
 & DiffPure & 0.3887  & 0.6976 & 0.4683  & 0.1342  & 0.4996  & 0.7364  & 0.6279  & 0.1272  & 0.5247  & 0.7021  & 0.6064  & 0.1340 \tabularnewline
 & DDA & 0.3838  & 0.7083  & 0.4776  & 0.1344  & 0.5337  & 0.7655  & 0.6544  & 0.1251  & 0.5105  & 0.7078  & 0.5875 & 0.1311\tabularnewline
 & \textsc{CosalPure} & \topone{0.4659}  & \topone{0.7327} & \topone{0.5487}  & \topone{0.1221}  & \topone{0.5946}  & \topone{0.7999}  & \topone{0.6881}  & \topone{0.1144}  & \topone{0.5859}  & \topone{0.7432} & \topone{0.6605}  & \topone{0.1215} \tabularnewline
\hline 
\hline 
\multirow{4}{*}{\rotatebox{90}{CoCA}} & Source-Only & 0.1837  & 0.5490 & 0.3402  & 0.1168  & 0.2339  & 0.5177  & 0.4698  & 0.1227  & 0.2239  & 0.4296  & 0.4082 & \topone{0.1500} \tabularnewline
 & DiffPure & 0.1706  & 0.5362  & 0.3492  & 0.1213 & 0.2231  & 0.5051  & 0.4995  & 0.1190  & 0.2185  & 0.4426  & 0.4286  & 0.1649 \tabularnewline
 & DDA & 0.2054 & 0.5543 & 0.3668  & 0.1156  & 0.2671 & 0.5476  & 0.5165  & 0.1129  & 0.2416  & 0.4503  & 0.4470  & 0.1548 \tabularnewline
 & \textsc{CosalPure} & \topone{0.2409}  & \topone{0.5753}  & \topone{0.3976}  & \topone{0.1119}  & \topone{0.3057}  & \topone{0.5884}  & \topone{0.5512} & \topone{0.1040}  & \topone{0.2633}  & \topone{0.4681}  & \topone{0.4745}  & 0.1604 \tabularnewline
\hline 
\end{tabular}

}
\end{table*}
\begin{table*}[ht]
\caption{Co-saliency detection success rates (SR) of entire group of images, only adversarial images and only clean images.
This table is to intuitively illustrate the impact of different methods on the adversarial and clean portions of group images.}
\vspace{-5pt}
    \centering
    \label{Table:baselines_split}
    \setlength{\tabcolsep}{12pt}
    \resizebox{1\linewidth}{!}{

\begin{tabular}{ll|ccc|ccc|ccc}
\hline 
\multirow{2}{*}{} & \multirow{2}{*}{} & \multicolumn{3}{c|}{GICD} & \multicolumn{3}{c|}{GCAGC} & \multicolumn{3}{c}{PoolNet}\tabularnewline
 &  & avg $\uparrow$ & adv $\uparrow$ & clean $\uparrow$ & avg $\uparrow$ & adv $\uparrow$ & clean $\uparrow$ & avg $\uparrow$ & adv $\uparrow$ & clean $\uparrow$\tabularnewline
\hline 
\hline 
\multirow{4}{*}{\rotatebox{90}{Cosal2015}} & Source-Only & 0.3493 & 0.1053  & 0.5884  & 0.5285  & 0.3741  & 0.6797  & 0.5677  & 0.3671  & 0.7642 \tabularnewline
 & DiffPure & 0.4595  & 0.3560  & 0.5609  & 0.4987  & 0.4533  & 0.5432  & 0.6327  & 0.5636  & 0.7003 \tabularnewline
 & DDA & 0.4565  & 0.3079  & 0.6021  & 0.5955  & 0.4924  & 0.6964  & 0.6233  & 0.5185  & 0.7259 \tabularnewline
 & \textsc{CosalPure} & \topone{0.5602}  & \topone{0.5416}  & 0.5785  & \topone{0.5975}  & \topone{0.5977}  & 0.5972  & \topone{0.6908}  & \topone{0.6569}  & 0.7239 \tabularnewline
\hline 
\hline 
\multirow{4}{*}{\rotatebox{90}{iCoseg}} & Source-Only & 0.4012 & 0.1516  & 0.6336  & 0.6469  & 0.6161  & 0.6756  & 0.5847  & 0.3451  & 0.8078 \tabularnewline
 & DiffPure & 0.4447  & 0.3161  & 0.5645  & 0.6609  & 0.6258  & 0.6936  & 0.6796  & 0.5741  & 0.7777 \tabularnewline
 & DDA & 0.4665  & 0.3483  & 0.5765  & 0.6982  & 0.6903  & 0.7057  & 0.6578  & 0.5419  & 0.7657 \tabularnewline
 & \textsc{CosalPure} & \topone{0.5396}  & \topone{0.5129}  & 0.5645  & \topone{0.7060}  & \topone{0.7064}  & 0.7057  & \topone{0.7278}  & \topone{0.6645}  & 0.7867\tabularnewline
\hline 
\hline 
\multirow{4}{*}{\rotatebox{90}{CoSOD3k}} & Source-Only & 0.3281  & 0.1118  & 0.5364  & 0.4445  & 0.2901  & 0.5932  & 0.4466  & 0.2354  & 0.6500 \tabularnewline
 & DiffPure & 0.3887  & 0.2987  & 0.4754  & 0.4996  & 0.4333 & 0.5636  & 0.5247  & 0.4462  & 0.6003\tabularnewline
 & DDA & 0.3838  & 0.2606  & 0.5026  & 0.5337  & 0.4394  & 0.6246  & 0.5105  & 0.3945  & 0.6222 \tabularnewline
 & \textsc{CosalPure} & \topone{0.4659}  & \topone{0.4597}  & 0.4718  & \topone{0.5946}  & \topone{0.5955}  & 0.5938 & \topone{0.5859}  & \topone{0.5703}  & 0.6009 \tabularnewline
\hline 
\hline 
\multirow{4}{*}{\rotatebox{90}{CoCA}} & Source-Only & 0.1837  & 0.0877  & 0.2739  & 0.2339  & 0.1818  & 0.2829  & 0.2239 & 0.1371  & 0.3053 \tabularnewline
 & DiffPure & 0.1706  & 0.1212  & 0.2170  & 0.2231  & 0.1802  & 0.2634  & 0.2185  & 0.1754  & 0.2589\tabularnewline
 & DDA & 0.2054  & 0.1499  & 0.2574 & 0.2671  & 0.2264  & 0.3053  & 0.2416 & 0.1897  & 0.2904 \tabularnewline
 & \textsc{CosalPure} & \topone{0.2409}  & \topone{0.2360}  & 0.2455  & \topone{0.3057}  & \topone{0.2998}  & 0.3113  & \topone{0.2633}  & \topone{0.2264}  & 0.2979\tabularnewline
\hline 
\end{tabular}

}
\vspace{-10pt}
\end{table*}

\section{Experiment}

\subsection{Experimental Setup}

\noindent\textbf{Datasets.}
We conduct experiments on Cosal2015 \cite{zhang2016detection}, iCoseg \cite{batra2010icoseg}, CoSOD3k \cite{fan2020taking}, and CoCA \cite{zhao2019egnet}. These four datasets contain 2,015, 643, 3,316, and 1,295 images of 50, 38, 160, and 80 groups respectively.
%
%
We apply the SOTA adversarial attack for CoSOD (\ie, Jadena \cite{gao2022can}) to the first $50\%$ of images in each group, while the remaining $50\%$ of images are kept in the clean state.
We select the ``augment'' version of Jadena and follow the settings\cite{gao2022can}.

\noindent\textbf{Evaluation settings.}
%
We choose GICD \cite{zhang2020gradient} and GCAGC \cite{zhang2020adaptive} to evaluate our method as they are commonly used state-of-the-art CoSOD methods. 
Additionally, we take PoolNet \cite{liu2019simple} into consideration, assessing the performance in salient object detection.
%
%

\begin{figure*}[t]
\centering
\includegraphics[width=0.6\linewidth]{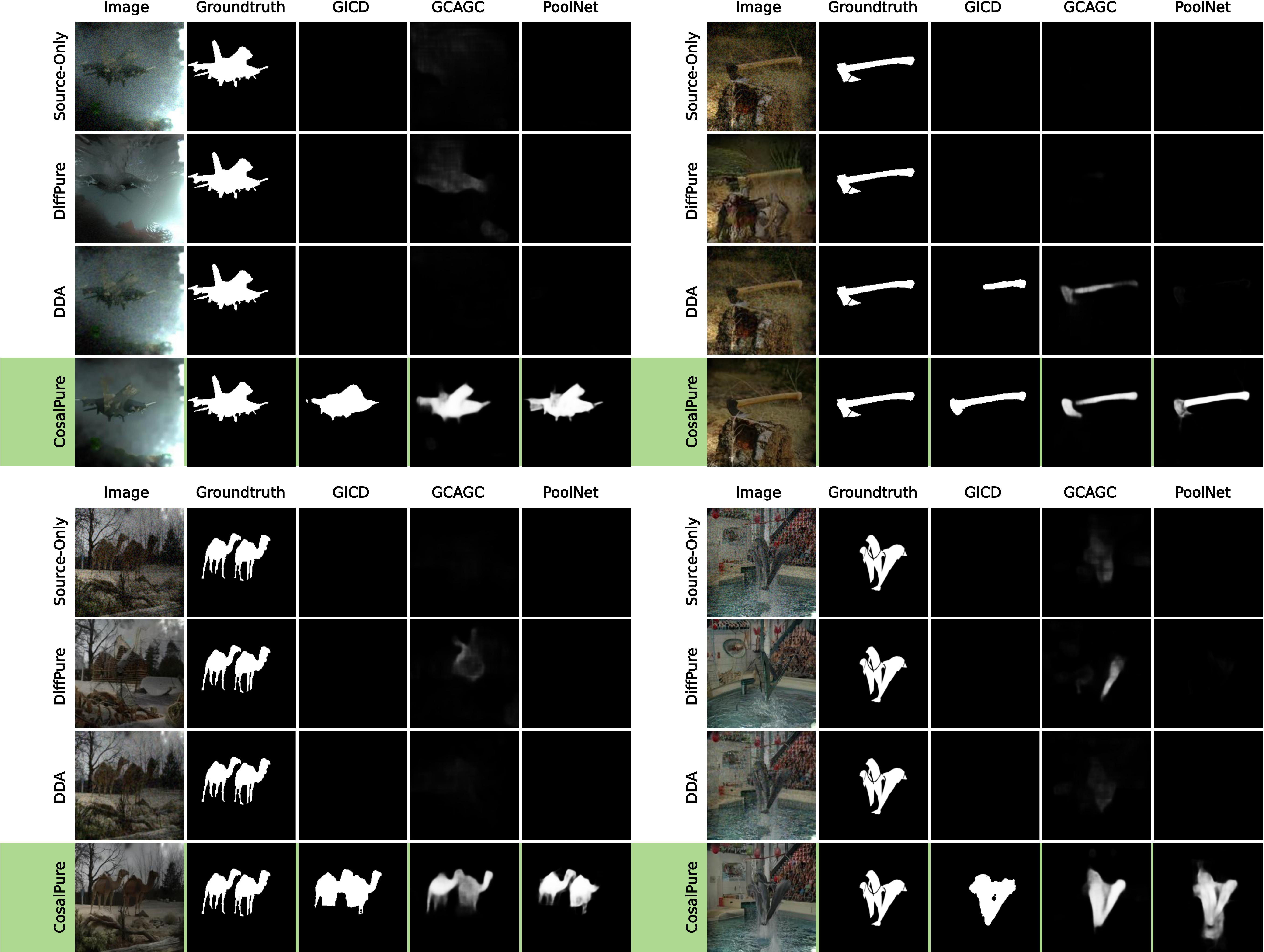}
\vspace{-5pt}
\caption{Visualization of co-salient object detection results. 
We show four visualized cases in this figure, with the source-only/purified image, the ground-truth of the co-saliency map, and the results of GICD, GCAGC, and PoolNet in the columns.
Our method, \textsc{CosalPure}, is highlighted in \textcolor[RGB]{174,217,144}{green}.
}
\label{fig:main_performance}
\vspace{-15pt}
\end{figure*}

\noindent\textbf{Baseline methods.}
Indeed, there is currently no specific image processing method designed for CoSOD attacks.
Hence, we employ two alternative approaches as baselines.
DiffPure \cite{nie2022diffusion} is a method that utilizes a diffusion model for purifying perturbation-based adversarial images.
Diffusion-Driven Adaptation (DDA) \cite{gao2022back} builds upon a diffusion-based model by introducing a novel self-ensembling scheme, enhancing the adaptation process by dynamically determining the degree of adaptation.
DiffPure and DDA employ the same sampling noise scale as our proposed \textsc{CosalPure}.

\noindent\textbf{Metrics.} 
We employ four metrics to evaluate the co-salient object detection result, including detection success rate (SR), average precision (AP) \cite{zhang2018review}, F-measure score $F_\beta$ with $\beta^2 = 0.3$ \cite{achanta2009frequency} and mean absolute error (MAE) \cite{zhang2018review}.
For the detection success rate, we calculate the intersection over union (IOU) between each co-salient object detection result of the reconstructed image and the corresponding ground-truth map.
We divide the number of successful results (IOU > 0.5) by the total number of results to calculate SR.
%
%
In addition, to intuitively illustrate the impact of different methods on CoSOD results, we not only compute SR for the entire group of images but also separately calculate SR for only adversarial images and only clean images. 


\noindent\textbf{Implementation details.} 
In the group-image concept learning procedure, the sampled images are simply resized from $224\times224$ resolution to $768\times768$ resolution before being passed into the image encoder.
For the continuous representation \cite{chen2021learning,ho2022disco} module employed in the concept-guided diffusion purification procedure, we constructed a dataset to train it.
We select 50,000 samples with a resolution of $224\times224$ from ImageNet (1,000 categories, each with 50 samples) and apply noise with the intensity of 16/255 via the PGD attack to these samples to construct the inputs of the continuous representation module.
For the ground truth images, we apply clean images with a resolution of $768\times768$ corresponding to the input images.
We follow the experimental setup of \cite{ho2022disco} and trained for 10 epochs to obtain the required module.
The group-image concept learning procedure and the concept-guided purification procedure utilize the same pre-trained image encoder and conditional diffusion model.
For the concept-guided purification procedure, we set the number of timesteps $T$ to 250, and the same configuration is applied to baseline methods.

\subsection{Comparison on Adversarial Attacks}

\begin{table*}[t]
\caption{Ablation study.
``w/o concept inversion'' represents only utilize the continuous representation module and not apply the subsequent purification process.
``w/ None concept'' denotes passing a meaningless ``None'' as the concept during the purification procedure.
``w/ learned concept'' denotes the complete pipeline, firstly learning the concept from the entire group of images and secondly passing in the learned concept during the purification procedure.}
\vspace{-5pt}
    \centering
    \label{Table:ablation}
    \setlength{\tabcolsep}{6pt}
    \resizebox{1\linewidth}{!}{

\begin{tabular}{ll|cccc|cccc|cccc}
\hline 
\multirow{2}{*}{} & \multirow{2}{*}{} & \multicolumn{4}{c|}{GICD} & \multicolumn{4}{c|}{GCAGC} & \multicolumn{4}{c}{PoolNet}\tabularnewline
 &  & SR $\uparrow$  & AP $\uparrow$ & $F_\beta$ $\uparrow$ & MAE $\downarrow$  &  SR $\uparrow$  & AP $\uparrow$ & $F_\beta$ $\uparrow$ & MAE $\downarrow$  &  SR $\uparrow$  & AP $\uparrow$ & $F_\beta$ $\uparrow$ & MAE $\downarrow$ \tabularnewline
\hline 
\hline 
\multirow{4}{*}{\rotatebox{90}{Cosal2015}} & Source-Only & 0.3493 & 0.7306 & 0.4038 & 0.1676 & 0.5285 & \topone{0.7853} & 0.6302 & \topone{0.1570} & 0.5677 & 0.7425 & 0.6095 & 0.1276\tabularnewline
 & \textsc{CosalPure} w/o concept inversion & 0.5186 & 0.7791 & 0.5809 & 0.1350 & 0.5528 & 0.7322 & 0.6430 & 0.2145 & 0.6843 & 0.8241 & 0.7225 & 0.1098\tabularnewline
 & \textsc{CosalPure} w/ None concept & 0.5225 & 0.7784 & 0.5886 & 0.1354 & 0.5334 & 0.7091 & 0.6132 & 0.2263 & 0.6774 & 0.8172 & 0.7160 & 0.1110\tabularnewline
 & \textsc{CosalPure} w/ learned concept & \topone{0.5602} & \topone{0.7898} & \topone{0.6177} & \topone{0.1296} & \topone{0.5975} & 0.7449 & \topone{0.6521} & 0.2063 & \topone{0.6908} & \topone{0.8268} & \topone{0.7258} & \topone{0.1086}\tabularnewline
\hline 
\hline 
\multirow{4}{*}{\rotatebox{90}{CoSOD3k}} & Source-Only & 0.3281 & 0.6988 & 0.4003 & 0.1439 & 0.4445 & 0.7325 & 0.5702 & 0.1376 & 0.4466 & 0.6606 & 0.5255 & 0.1386\tabularnewline
 & \textsc{CosalPure} w/o concept inversion & 0.4424 & 0.7314 & 0.5317 & 0.1243 & 0.5753 & 0.7923 & 0.6804 & 0.1170 & 0.5747 & 0.7424 & 0.6508 & 0.1223\tabularnewline
 & \textsc{CosalPure} w/ None concept & 0.4297 & 0.7216 & 0.5158 & 0.1273 & 0.5488 & 0.7715 & 0.6584 & 0.1288 & 0.5711 & 0.7373 & 0.6453 & 0.1231\tabularnewline
 & \textsc{CosalPure} w/ learned concept & \topone{0.4659} & \topone{0.7327} & \topone{0.5487} & \topone{0.1221} & \topone{0.5946} & \topone{0.7999} & \topone{0.6881} & \topone{0.1144} & \topone{0.5859} & \topone{0.7432} & \topone{0.6605} & \topone{0.1215}\tabularnewline
\hline 
\end{tabular}

}
\vspace{-10pt}
\end{table*}

We denote the images before reconstruction (containing 50\% adversarial images and 50\% clean images) as "Source-Only".
The comparison between our proposed \textsc{CosalPure} and baselines are shown in Table~\ref{Table:baselines}.
We consider an image to be successfully detected in the co-salient object detection (CoSOD) task if the IOU of its CoSOD result and the ground-truth map exceed 0.5.
Compared to DiffPure \cite{nie2022diffusion} and DDA \cite{gao2022back}, \textsc{CosalPure} outperforms them in terms of co-salient object detection success rates (SR) across all four datasets.
For the other three metrics (\ie, AP \cite{zhang2018review}, $F_\beta$ \cite{achanta2009frequency} and MAE \cite{zhang2018review}), \textsc{CosalPure} remains the best at most of the time.
We show four visualized cases in \figref{fig:main_performance}.
For each case, we present the generated images of \textsc{CosalPure} and two baselines, DiffPure and DDA.
Obviously \textsc{CosalPure} generates higher-quality images, as it leverages the intrinsic commonality of objects across the group of images.
Additionally, we showcase the comparison of the detection results on GICD, GCAGC, and PoolNet.
The results from \textsc{CosalPure} closely approximate the ground-truth map, while the baseline methods struggle to display the correct results.

To intuitively illustrate the impact of different methods on the adversarial and clean portions of group images, we measure the co-salient object detection success rate from three perspectives.
In Table~\ref{Table:baselines_split}, "avg" represents the evaluation across the entire group of images, "adv" and "clean" correspond to evaluations on only the 50\% images that are under the SOTA attack \cite{gao2022can} and on only the 50\% images that remain clean.
\textsc{CosalPure} at some times have a lower ``clean'' SR compared to DDA or source-only.
However, DiffPure and DDA are unsatisfactory in ``adv'' SR, while \textsc{CosalPure} exhibits a significant lead in ``adv'' SR, resulting in it consistently performing the best in ``avg'' SR.

\subsection{Ablation Study}

\begin{figure}[t]
\centering
\includegraphics[width=1\linewidth]{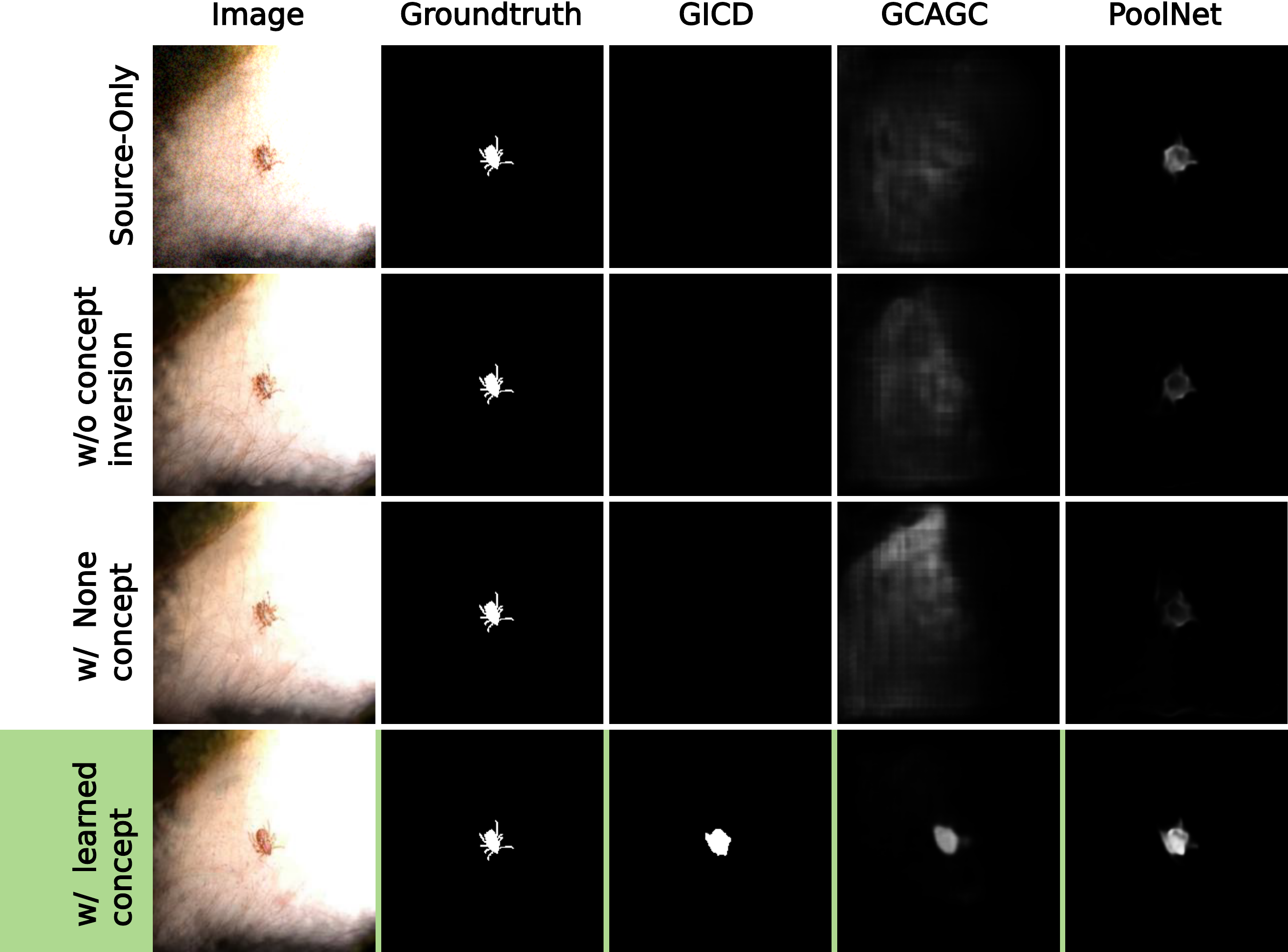}
\vspace{-15pt}
\caption{Visualization for ablation study.}
\label{fig:ablation}
\vspace{-5pt}
\end{figure}
To validate the effect of the learned concepts on CoSOD results, we conduct ablation studies on Cosal2015 \cite{zhang2016detection} and CoSOD3k \cite{fan2020taking}.
In Table~\ref{Table:ablation}, ``w/o concept inversion'' represents only utilizing the continuous representation module and not applying the subsequent purification process.
``w/ None concept'' denotes passing a meaningless ``None'' as the concept during the purification.
``w/ learned concept'' denotes the complete pipeline, firstly learning the concept from the entire group of images and secondly passing the learned concept during the purification procedure to accomplish image reconstruction.
From Table~\ref{Table:ablation}, it is evident that the learned concept contributes to significant improvements in various metrics.
As shown in \figref{fig:ablation}, when we do not apply the concept-guided purification or when we pass in a meaningless concept, the generated image performs poorly in the CoSOD task.
This improves when we pass in the learned effective concept which includes object semantics.
The example proves that the learned concept contributes to the reconstruction of images used for the CoSOD task.

\subsection{Extention to Common Corruption}

\begin{table}[t]
\caption{Extension to motion blur.}
\vspace{-5pt}
    \centering
    \label{Table:motion_blur}
    \setlength{\tabcolsep}{12pt}
    \resizebox{1\linewidth}{!}{

\begin{tabular}{l|cccc}
\hline 
\multirow{1}{*}{} & SR $\uparrow$  & AP $\uparrow$ & $F_\beta$ $\uparrow$ & MAE $\downarrow$\tabularnewline
\hline 
Source-Only & 0.3915 & 0.7408 & 0.4373 & 0.1590\tabularnewline
DiffPure & 0.3146 & 0.6774 & 0.3763 & 0.1738\tabularnewline
DDA & 0.3900 & 0.7381 & 0.4425 & 0.1579\tabularnewline
\textsc{CosalPure} & \topone{0.4575} & \topone{0.7419} & \topone{0.5241} & \topone{0.1505}\tabularnewline
\hline 
\end{tabular}

}
\vspace{-15pt}
\end{table}

In addition to adversarial attacks on CoSOD, we also broaden our experiments to include a common corruption type: motion blur. 
We select the Cosal2015 dataset \cite{zhang2016detection} and, similar to the adversarial experiments, apply motion blur \cite{hendrycks2019robustness} to the first 50\% of images in each group while keeping the remaining 50\% of images clean.
Here we set the number of timesteps $T$ to 500 and do not employ the continuous representation module.
As shown in Table~\ref{Table:motion_blur}, \textsc{CosalPure} performs better than other diffusion-based image processing methods at all four metrics.


\section{Conclusions}
\label{sec:conclu}

This paper presented \textsc{CosalPure}, an innovative framework enhancing the robustness of co-salient object detection (CoSOD) against adversarial attacks and common image corruptions. Central to our approach are two key innovations: group-image concept learning and concept-guided diffusion purification. Our framework effectively captures and utilizes the high-level semantic concept of co-salient objects from group images, demonstrating notable resilience even in the presence of adversarial examples.

Empirical evaluations across datasets like Cosal2015, iCoseg, CoSOD3k, and CoCA showed that \textsc{CosalPure} significantly outperforms existing methods such as DiffPure and DDA in CoSOD tasks. Not only did it achieve higher success rates, but it also excelled in performance metrics like AP, F-measure, and MAE. Additionally, its effectiveness against common image corruptions, like motion blur, underscores its versatility.

Our \textsc{CosalPure} represents a substantial advancement in CoSOD, offering robust, concept-driven image purification. It opens avenues for more resilient co-salient object detection, vital in today's landscape of sophisticated image manipulation and corruption. Future work might extend this framework to broader image analysis applications and explore its adaptability to real-world scenarios.

\section{Acknowledgments}
Geguang Pu is supported by National Key Research and Development Program (2020AAA0107800), and Shanghai Collaborative Innovation Center of  Trusted Industry Internet Software. This work is also supported by the National Research Foundation, Singapore, and DSO National Laboratories under the AI Singapore Programme (AISG Award No: AISG2-GC-2023-008), and Career Development Fund (CDF) of the Agency for Science, Technology and Research (A*STAR) (No.: C233312028).

\clearpage
\newpage

{
    \small
    \bibliographystyle{ieeenat_fullname}
    \bibliography{ref}
}


\end{document}


\title{Supplementary Materials for ACM Multimedia 2023 paper \#1196}

\author{First Author\\
Institution1\\
Institution1 address\\
{\tt\small firstauthor@i1.org}
\and
Second Author\\
Institution2\\
First line of institution2 address\\
{\tt\small secondauthor@i2.org}
}

\maketitle

This paper is the supplementary material for ACM Multimedia 2023 paper \#1196, including the details of loss functions and the realistic validation of our joint deraining~\&~deblurring dataset (\textit{JD$^3$}).

\section{Loss Functions}

We consider two loss functions for training our proposed MARIN.
To ensure that the restoration image is close to the ground truth in content, we use the multi-scale content loss function \cite{nah2017deep} as the first part loss function, formulated as
\begin{align}
    L_{cont} = \sum_{l=1}^{L} {\| \mathbf{I}' - \mathbf{I}^* \|}_1,
    \label{eq:content_loss}
\end{align}
where $L$ denotes the number of scales, \ie, the layers number of the encoder.
$\mathbf{I}'$ and $\mathbf{I}^*$ represent the resulting image processed by MARIN and the sharp rain-free image as the ground truth at the $l$-th scale, respectively.

In a recent study \cite{jiang2021focal}, minimizing the distance between the generated image and the ground truth in the frequency space is effective in image restoration tasks.
To this end, we apply a multi-scale frequency loss function as follows:
\begin{align}
    L_{freq} = \sum_{l=1}^{L} {\| \mathbb{F}(\mathbf{I}') - \mathbb{F}(\mathbf{I}^*) \|}_1,
    \label{eq:frequency_loss}
\end{align}
where $\mathbb{F}$ represents the fast Fourier transform function that calculates the frequency domain from an image. Finally, we set a ratio $\eta$ to combine the multi-scale content loss and the multi-scale frequency loss as
\begin{align}
    L_{whole} = L_{cont} + \eta \cdot L_{freq}.
    \label{eq:whole_loss}
\end{align}

\section{Realistic validation}
We conducted user studies to validate the realistic of our joint deraining~\&~deblurring dataset (\textit{JD$^3$}) following \cite{halder2019physics}.
As there was no existing similar dataset in this joint task, we separately conducted two user studies for rain realism and blur realism.
Our user studies involved 15 participants (8 males and 7 females, all graduate students), they are asked to make a rating (from strongly agree, agree, neutral, disagree and strongly disagree) for the realism of each image.
For each dataset, we randomly selected 20 images, so we recevie 300 ratings in total per dataset.

\vspace{5pt}
\noindent\textbf{Rain realism.}
We compare the rating result of our proposed \textit{JD$^3$} with Rain100H \cite{yang2017deep}, Rain1400 \cite{fu2017clearing}, and PbRain \cite{halder2019physics}, as they are widely used rain datasets.
The rating results for rain realism are shown in \figref{fig:bar_rain}.
The realism of rain streaks of \textit{JD$^3$} far outweighs Rain100H and Rain1400, slightly below the state-of-the-art rain dataset PbRain.

\vspace{5pt}
\noindent\textbf{Blur realism.}
For blur realism, we chose GOPRO \cite{nah2017deep} and REDS \cite{nah2019ntire} as compared datasets.
During to the similar synthetic process (\ie, averaging from multiple sharp frames captured by high-frame-rate cameras), the rating result of REDS, GOPRO and \textit{JD$^3$} are close. 
As the realistic property of REDS and GOPRO has been validated and widely used in the deblurring task, we judge \textit{JD$^3$} is realistic enough.

\begin{figure}
\centering
\includegraphics[width=1\linewidth]{image/bar_rain_crop.pdf}
\caption{
User study of rain realism.
}
\label{fig:bar_rain}
\end{figure}

\begin{figure}
\centering
\includegraphics[width=1\linewidth]{image/bar_blur_crop.pdf}
\caption{
User study of blur realism.
}
\label{fig:bar_blur}
\end{figure}

	\bibliographystyle{ACM-Reference-Format}
	\bibliography{acmart.bib}